\documentclass[letterpaper, 10 pt, conference]{ieeeconf}  

\IEEEoverridecommandlockouts                              

\usepackage{amsmath,amssymb,amsfonts}
\usepackage{algorithm, algorithmicx}
\usepackage{booktabs}
\usepackage[noend]{algpseudocode}
\usepackage{kotex}
\usepackage{graphicx}
\usepackage{multirow}
\usepackage{svg}
\usepackage{hyperref}

\title{\LARGE \bf
Reinforcement Learning-based Fault-Tolerant Control for Quadrotor with Online Transformer Adaptation}

\author{Dohyun Kim$^{1}$ \href{https://orcid.org/0009-0006-9338-4630}{\includegraphics[scale=0.5]{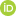}}, Jayden Dongwoo Lee$^{2}$ \href{https://orcid.org/0000-0002-6204-0644}{\includegraphics[scale=0.5]{figure/ORCID-iD_icon_16x16.png}}, Hyochoong Bang$^{2}$ \href{https://orcid.org/0000-0001-6016-8102}{\includegraphics[scale=0.5]{figure/ORCID-iD_icon_16x16.png}} and Jungho Bae$^{1}$ \href{https://orcid.org/0009-0005-1031-2694}{\includegraphics[scale=0.5]{figure/ORCID-iD_icon_16x16.png}}
\thanks{This work was supported by the Agency for Defense Development Grant funded by the Korean Government in 2025.}
\thanks{$^{1}$Dohyun Kim and Jungho Bae are with the Agency for Defense Development, Republic of Korea {\tt\small \{00dh.kim, daewith\}@gmail.com}}%
\thanks{$^{2}$Jayden Dongwoo Lee and Hyochoong Bang are with the Korea Advanced Institute of Science and Technology, Republic of Korea {\tt\small \{cin6474, hcbang\}@kaist.ac.kr}}%
}

\begin{document}

\maketitle
\thispagestyle{empty}
\pagestyle{empty}

\begin{abstract}

Multirotors play a significant role in diverse field robotics applications but remain highly susceptible to actuator failures, leading to rapid instability and compromised mission reliability. While various fault-tolerant control (FTC) strategies using reinforcement learning (RL) have been widely explored, most previous approaches require prior knowledge of the multirotor model or struggle to adapt to new configurations. To address these limitations, we propose a novel hybrid RL-based FTC framework integrated with a transformer-based online adaptation module. Our framework leverages a transformer architecture to infer latent representations in real time, enabling adaptation to previously unseen system models without retraining. We evaluate our method in a PyBullet simulation under loss-of-effectiveness actuator faults, achieving a 95\% success rate and a positional root mean square error (RMSE) of 0.129 m, outperforming existing adaptation methods with 86\% success and an RMSE of 0.153 m. Further evaluations on quadrotors with varying configurations confirm the robustness of our framework across untrained dynamics. These results demonstrate the potential of our framework to enhance the adaptability and reliability of multirotors, enabling efficient fault management in dynamic and uncertain environments. Website is available at \href{http://00dhkim.me/paper/rl-ftc}{http://00dhkim.me/paper/rl-ftc}

\end{abstract}

\section{INTRODUCTION}

Multirotors have emerged as effective tools in diverse field robotics applications, including logistics, surveillance, and agriculture.
Stable and robust control of multirotors in challenging environments remains critical.
In particular, multirotors generate lift through motor rotation, making them especially vulnerable to actuator faults such as loss-of-effectiveness (LoE) in certain motors.
These faults immediately cause instability, making multirotors more vulnerable than fixed-wing aircraft.
Therefore, actuator faults are critical in multirotor flight, requiring fault-tolerant control (FTC) strategies to ensure reliability.

Recently, FTC methods using reinforcement learning (RL) for multirotors have received attention due to their potential to learn complex and flexible control strategies.
Existing studies have successfully demonstrated RL's capability to maintain system performance under various fault conditions, such as torque distribution \cite{deng2023fault} and underwater propeller faults \cite{qin2024fault}, and fixed-wing UAV control \cite{giral2024transformer}.
Extensive research has particularly explored RL-based FTC for multirotors.
For instance, a parallel combination of a model-free RL-based FTC and original controllers was proposed as a countermeasure against UAV signal spoofing~\cite{fei2020learn}.
An RL-based policy that integrates outputs from both high-gain and low-gain PID controllers was developed~\cite{sohege2021novel}.
A method for generating FTC signals by leveraging discrepancies between fault-affected sensed states and states from a predefined health model is introduced~\cite{liu2024reinforcement}.

However, these studies face limitations when deployed in practical field scenarios. They often require prior system-specific health models or extensive retraining when encountering previously unseen dynamics, such as mid-mission changes in payload mass or variations in multirotor configuration.
Such constraints significantly reduce the adaptability and flexibility required in realistic field robotics applications.

To address these challenges, we propose an RL-based hybrid FTC framework combined with an adaptation module inspired by the rapid motor adaptation~(RMA) method~\cite{kumar2021rma}.
Unlike previous approaches, our approach employs an adaptation module that adopts a control strategy suitable for the current drone configuration based on sensor data and command history.
This method enables real-time, online adaptation without requiring additional training, thus facilitating adaptation to previously unseen dynamics, including mid-flight changes in mass or inertia.
The adaptation module supports policies in adapting to diverse and uncertain environments.
The RMA module developed by \cite{kumar2021rma} has demonstrated successful real-time adaptation in quadruped robot control tasks.
Extending from ground robots to aerial systems, recent studies by \cite{zhang2023learning, zhang2024learning} have achieved online adaptation for quadrotors, demonstrating robustness against environmental variations without additional training.
Furthermore, \cite{hsu2024reforma} successfully adapted quadrotor control systems to external disturbances.
While these studies emphasize environmental adaptation and resilience to minor oscillation, our work uniquely addresses significant actuator faults, such as substantial thrust losses reaching up to 36\%.

Further enhancing our framework, we propose a novel transformer-based adaptation module to enhance domain inference capabilities beyond existing approaches.
We validate our method in simulated hovering tasks using PyBullet~\cite{pybullet}, demonstrating robust and reliable position control despite substantial actuator faults.
Our results emphasize the practical applicability of our proposed framework, advancing the robustness and flexibility of autonomous aerial systems.

\section{METHOD}
\begin{figure}[t!]
  \centering
  \includegraphics[width=\columnwidth]{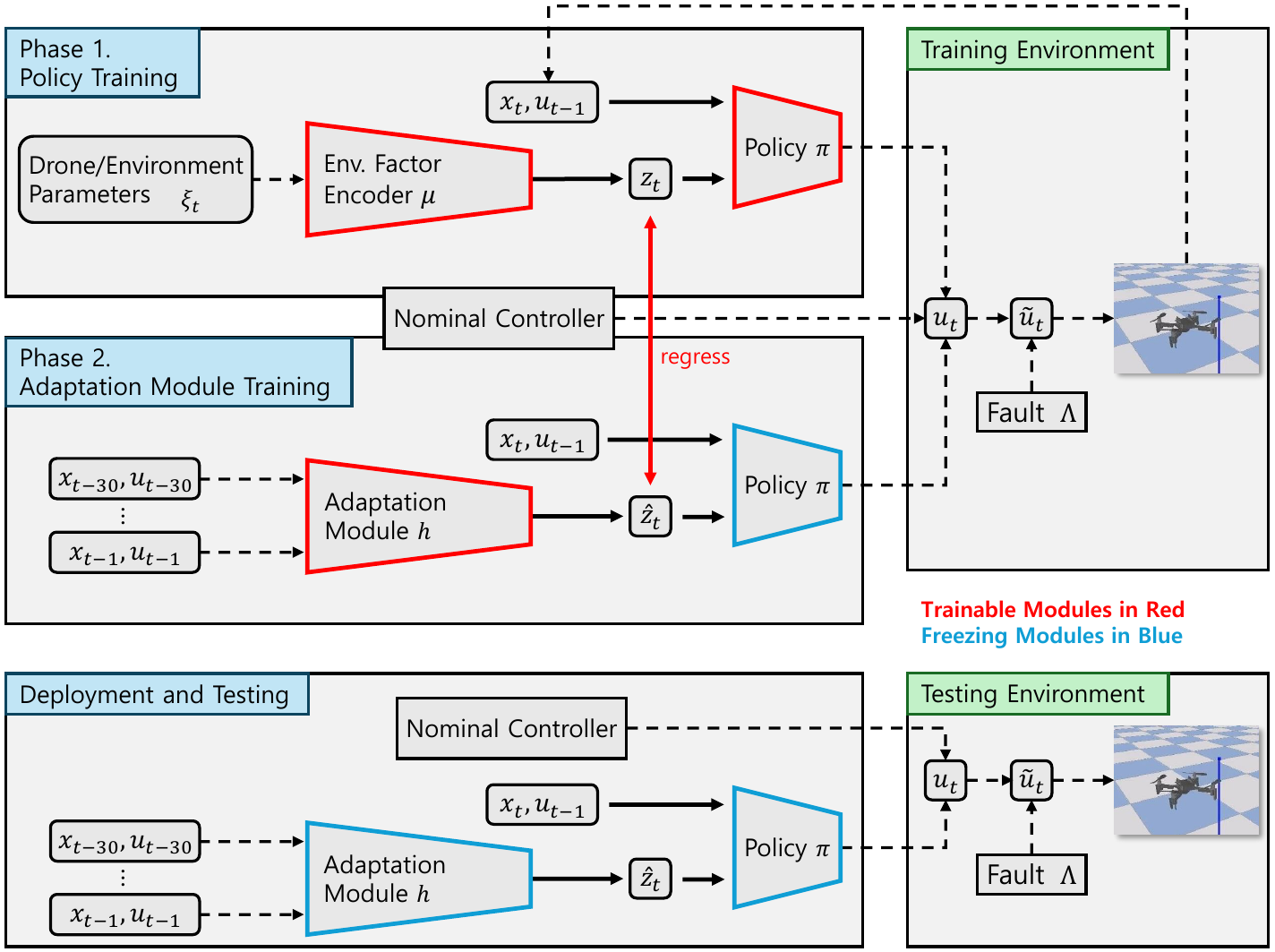}
  \caption{
  The overview of 2-phase training and deployment.
  \textbf{Policy Training:}
  In phase 1, the environmental factor encoder~$\mu$ and policy~$\pi$ are jointly trained.
  $\mu$ encodes privileged factors into a latent vector~$z_t$. The policy then uses $z_t$ together with state~$x_t$ and previous action~$u_{t-1}$ to generate the control command~$u^\text{rl}_t$.
  The command is added to the nominal controller's command~$u^\text{nominal}_t$ and applied to the dynamics model.
  \textbf{Adaptation Module Training:}
  The adaptation module~$h$ learns to infer $\hat z_t$ without access to privileged state~$\xi_t$, through supervised learning to minimize the error between $\hat z_t$ and ground truth~$z_t$.
  Training data is collected using the frozen policy $\pi$.
  \textbf{Deployment and Testing:}
  In deployment, both $\pi$ and $h$ are frozen.
  $h$ infers $\hat z_t$ from trajectory, which $\pi$ uses to produce control commands.
  This enables adaptability to unseen dynamics while achieving FTC.
  }
  \label{architecture}
\end{figure}

Our framework consists of a nominal controller, a policy, and an adaptation module.
This section describes the problem formulation, the 2-phase training process, and adaptation module.
For details, please see Fig.~\ref{architecture}.

\subsection{Problem Formulation}

\vspace{0.6em}\noindent\textbf{UAV Model Description:}\quad
We use the physics-based gym-pybullet-drones environment~\cite{panerati2021learning} to simulate multirotor dynamics, including attitude and position changes.
The thrust by each motor and torque induced around the multirotor's z-axis are expressed as follows:
\begin{equation}
F_{i} = k_F \cdot u_{t,i}^2, \quad 
T = \sum_{i=1}^{4} (-1)^{i} k_T \cdot u_{t,i}^2,
\end{equation}
where \( u_t = [u_{t,1}, u_{t,2}, u_{t,3}, u_{t,4}]^\intercal \) represents the revolutions per minute (RPM) of each motor, and \( k_F \) and \( k_T \) are the lift and torque coefficients, respectively.

\vspace{0.6em}\noindent\textbf{Fault Representation:}\quad
Actuators are subject to wear, propeller damage, and motor overheating.
These faults are modeled as a loss-of-effectiveness, reducing the RPM of certain motors.
The fault matrix $\Lambda$ is defined as:
\begin{equation}
\Lambda = \text{diag}(\lambda_1, \lambda_2, \lambda_3, \lambda_4), \quad
\tilde{u}_t=\Lambda u_t,
\end{equation}
where \(\lambda_i\) represents the efficiency of the \(i\)-th motor and is nominally 1.
The RPM with the fault is denoted as \(\tilde{u}_t\).
We consider faults affecting one motor with $\lambda_i$ between 0.8 to 0.9, leading to up to 36\% thrust loss.
This reflects realistic fault scenarios~\cite{liu2024reinforcement}.
Complete failure was not considered because this study focuses on partial damage situations.

\subsection{Hybrid Controller}
We propose a hybrid FTC controller for hovering under LoE actuator faults, combining a nominal controller and policy $\pi$.
We also introduce an environmental factor encoder $\mu$ as a teacher model during adaptation module training.

\vspace{0.6em}\noindent\textbf{Nominal Controller:}\quad
The nominal controller uses a cascade Proportional-Integral-Differential~(PID) structure for position and attitude control.
The outer-loop PID computes z-axis thrust command and desired attitude from position errors, while the inner-loop PID generates torque commands from attitude errors.
These commands are converted to RPM signals.
These PID gains are optimized for accurate tracking under fault-free conditions and are used in both training and testing.

\vspace{0.6em}\noindent\textbf{RL Policy:}\quad
The policy \(\pi\) assists the nominal controller for stable hovering under actuator faults.
At time \(t\), it takes the state \(x_t\), previous control input \(u_{t-1}\), and latent vector \(z_t\) to output motor commands \(u_t\).
The state \(x_t\) includes position, attitude, linear velocity and angular velocity.
The policy's control is capped at 20\% of the nominal command, ensuring a supportive role in fault handling.
\(z_t\), extracted from the environmental encoder or adaptation module, allows for adaptive control.
The control frequency is 30 Hz.

\vspace{0.6em}\noindent\textbf{Latent Representation:}\quad
The environmental factor encoder $\mu$ compresses privileged parameters $\xi_t$ that represent the dynamic characteristics, allowing the policy to maintain consistent control performance across various domains.
This approach aligns with previous research using domain adaptation methods \cite{kumar2021rma, zhang2023learning, zhang2024learning} and effectively captures environmental variations.
In addition, domain randomization is applied to facilitate generalization of both policy $\pi$ and encoder $\mu$.
Specifically, $\xi_t$ are randomly varied within $\pm 20\%$ to prevent the policy from overfitting to specific environments.

\vspace{0.6em}\noindent\textbf{Reward Function:}\quad
The reward function is designed to ensure stability despite faults.
It minimizes position and attitude errors, penalizes rapid control fluctuations, and promotes progressive success.
Let $\mathbf{p}$, $\mathbf{v}$, $\boldsymbol{\theta}$, and $\mathbf{p}_{\text{target}}$ denote the position, velocity, and attitude, and target position of multirotor, respectively.
The reward at time $t$ is defined as
\begin{align}
R_t = &-k_{\text{pos}} \|\mathbf{p} - \mathbf{p}_{\text{target}}\| 
      -k_{\text{att}} \|\boldsymbol{\theta}\| 
      -k_{\text{vel}} \|\mathbf{v}\|^2 \notag \\
      &-k_{\text{rate}} \|\dot{u}_t\|^2 
      -k_{\text{smooth}} \|\ddot{u}_t\|^2 
      + 0.001 
      + R_{\text{succ}} \\[2ex]
R_{\text{succ}} = &
\begin{cases}
    1.0, & \text{if } \|\mathbf{p} - \mathbf{p}_{\text{target}}\| < \epsilon_{\text{goal}} \text{ for 1 sec.} \\
    0.1, & \text{if } \|\mathbf{p} - \mathbf{p}_{\text{target}}\| < \epsilon_{\text{goal}} \\
    0.05, & \text{if } \|\mathbf{p} - \mathbf{p}_{\text{target}}\| < \epsilon_{\text{near}} \\
    0, & \text{otherwise,}
\end{cases}
\end{align}
where the scaling factors are set to 0.5, 0.1, 0.001, 0.5, and 0.1, respectively, and the thresholds are set to \(\epsilon_{\text{goal}} = 0.173\) and \(\epsilon_{\text{near}} = 0.520\).

\subsection{Transformer-based Adaptation Module}

Measuring quadrotor and environmental parameters $\xi_t$ in the real world is challenging.
To address this, we replace the encoder $\mu$ with an adaptation module.
To the best of our knowledge, this paper is the first application of RMA-style adaptation module for fault-tolerant control, enabling robust adaptive control without privileged information.
Our adaptation module $h$ infers a latent vector $\hat z_t$ from 1 second state-action trajectories.
The module is trained using a teacher-student paradigm.
The frozen environmental encoder $\mu$ acts as the teacher without training, and $h$ learns to mimic its output by minimizing the mean squared error $\|\hat z_t - z_t\|^2$.
Training data is collected using the frozen policy learned in Phase 1.

Unlike RMA~\cite{kumar2021rma}, we introduce a transformer architecture~\cite{vaswani2017attention} to improve inference performance.
Transformers leverage self-attention mechanisms to capture temporal correlations throughout the entire input sequence.
This approach effectively filters noise and vibration, and discriminates key physical characteristics.
This facilitates the stability of latent representations, ensuring robust inference performance even in unseen dynamics during training.

\section{EXPERIMENTS}

\subsection{Experiment Setup}
\begin{table}[tb]
  \centering
  \caption{Ranges of the physical parameters and configuration.}
  \label{tab:cf2_parameters}
  \begin{tabular}{l@{\hspace{4pt}}ccc}
    \toprule
    \multirow{2}{*}{\textbf{Parameter}} & \multirow{2}{*}{\textbf{CF}} & \multicolumn{2}{c}{\textbf{Test}} \\
                      &                & \textbf{UAV1} & \textbf{UAV2} \\
    \midrule
    Mass \scriptsize{(g)}                                  & 27            & 40.5          & 18.9 \\
    Arm Length \scriptsize{(m)}                            & 0.0397        & 0.0594        & 0.0397 \\
    Thrust Coef. \scriptsize{(N/(rad/s)$^2$, \(\times10^{-10}\))}                        &3.16      & 4.74      & 2.21 \\
    Torque Coef. \scriptsize{(N·m/(rad/s)$^2$, \(\times10^{-12}\))}                       & 7.94      & 1.19      & 5.56 \\
    Thrust-to-Weight Ratio                    & 2.25          & 2             & 2.8 \\
    Inertia \(I_{xx}\) \scriptsize{(g·m\(^2\), \(\times10^{-2}\))}            & 2.3951      & 4.73        & 0.98 \\
    Inertia \(I_{yy}\) \scriptsize{(g·m\(^2\), \(\times10^{-2}\))}            & 2.3951      & 4.73        & 0.98 \\
    Inertia \(I_{zz}\) \scriptsize{(g·m\(^2\), \(\times10^{-2}\))}            & 3.2347      & 7.32        & 1.52 \\
    \midrule
    Initial Position(X, Y) \scriptsize{(m)}                & \([-0.2,\,0.2]\)   & \multicolumn{2}{c}{\([-0.6,\,0.6]\)} \\
    Initial Position(Z) \scriptsize{(m)}                   & \([0.8,\,1.2]\)    & \multicolumn{2}{c}{\([0.7,\,1.3]\)} \\
    Fault Range                               & \([0.1,\,0.2]\)    & \multicolumn{2}{c}{\([0.1,\,0.2]\)} \\
    \bottomrule
  \end{tabular}%
\end{table}

\begin{table}[t]
    \centering
    \caption{Performance of our method and baselines for each quadrotor under 30\% LoE of $F_2$.}
    \label{tab:performance_and_std_dev}
    \resizebox{\columnwidth}{!}{%
    \begin{tabular}{c lccc}
        \toprule
        Model & Method & Success & RMSE \scriptsize{(m)} & Max Error \scriptsize{(m)} \\
        \midrule
        \multirow{4}{*}{CF}%
              & \textbf{RL-ours} & \textbf{95\%} & \textbf{0.129 ± 0.042} & \textbf{0.545 ± 0.135} \\
              & RL-RMA \cite{kumar2021rma} & 86\% & 0.153 ± 0.092 & 0.565 ± 0.145 \\
              & PID & 53\% & 1.220 ± 1.605 & 2.230 ± 2.178 \\
              & No Fault (oracle) & 100\% & 0.063 ± 0.015 & 0.486 ± 0.160 \\
        \midrule
        \multirow{4}{*}{UAV1}%
              & \textbf{RL-ours} & \textbf{91\%} & \textbf{0.165 ± 0.048} & \textbf{0.589 ± 0.132} \\
              & RL-RMA \cite{kumar2021rma} & 85\% & 0.182 ± 0.092 & 0.615 ± 0.149 \\
              & PID & 55\% & 1.340 ± 1.629 & 2.460 ± 2.231 \\
              & No Fault (oracle) & 100\% & 0.074 ± 0.013 & 0.487 ± 0.159 \\
        \midrule
        \multirow{4}{*}{UAV2}%
              & \textbf{RL-ours} & \textbf{96\%} & \textbf{0.109 ± 0.042} & \textbf{0.534 ± 0.150} \\
              & RL-RMA \cite{kumar2021rma} & 87\% & 0.136 ± 0.093 & 0.554 ± 0.156 \\
              & PID & 45\% & 1.286 ± 1.802 & 2.294 ± 2.587 \\
              & No Fault (oracle) & 100\% & 0.057 ± 0.016 & 0.486 ± 0.161 \\
        \bottomrule
    \end{tabular}%
    }
\end{table}

We evaluated our framework in gym-pybullet-drones~\cite{panerati2021learning}, a high-fidelity simulator.
The simulation used the Bitcraze Crazyflie 2.x (referred to as \textbf{CF}) as the base quadrotor, and evaluated adaptability with two variants: a larger and heavier \textbf{UAV1} and a lighter \textbf{UAV2}.
Their physical parameters are in Table~\ref{tab:cf2_parameters}.

In Phase 1, the policy $\pi$ and environmental encoder $\mu$ were jointly trained with proximal policy optimization~\cite{schulman2017proximal} algorithm.
In Phase 2, the adaptation module $h$ was trained using the ADAM~\cite{kingma2014adam} optimizer.
The environmental encoder used a 2-layer MLP with a 64-dimensional hidden layer.
The policy followed the Stable Baselines3~\cite{raffin2021stable} settings.
To compare adaptation module structures, both CNN-based and transformer-based modules were designed.
The CNN model had a 3-layer 1-D CNN architecture with input/output channel numbers, kernel sizes, and strides set to $[32, 32, 4, 2]$, $[32, 32, 3, 1]$, and $[32, 32, 3, 1]$, respectively.
The transformer model was configured with a 16-dimensional input, 128 model dimension, sequence length 64, 2 attention heads, 2 encoder layers, a 128 feed-forward dimension, and an 8-dimensional latent output.

\subsection{Experiment Results}

\begin{figure}[t!]
    \centering
    \includegraphics[width=\columnwidth]{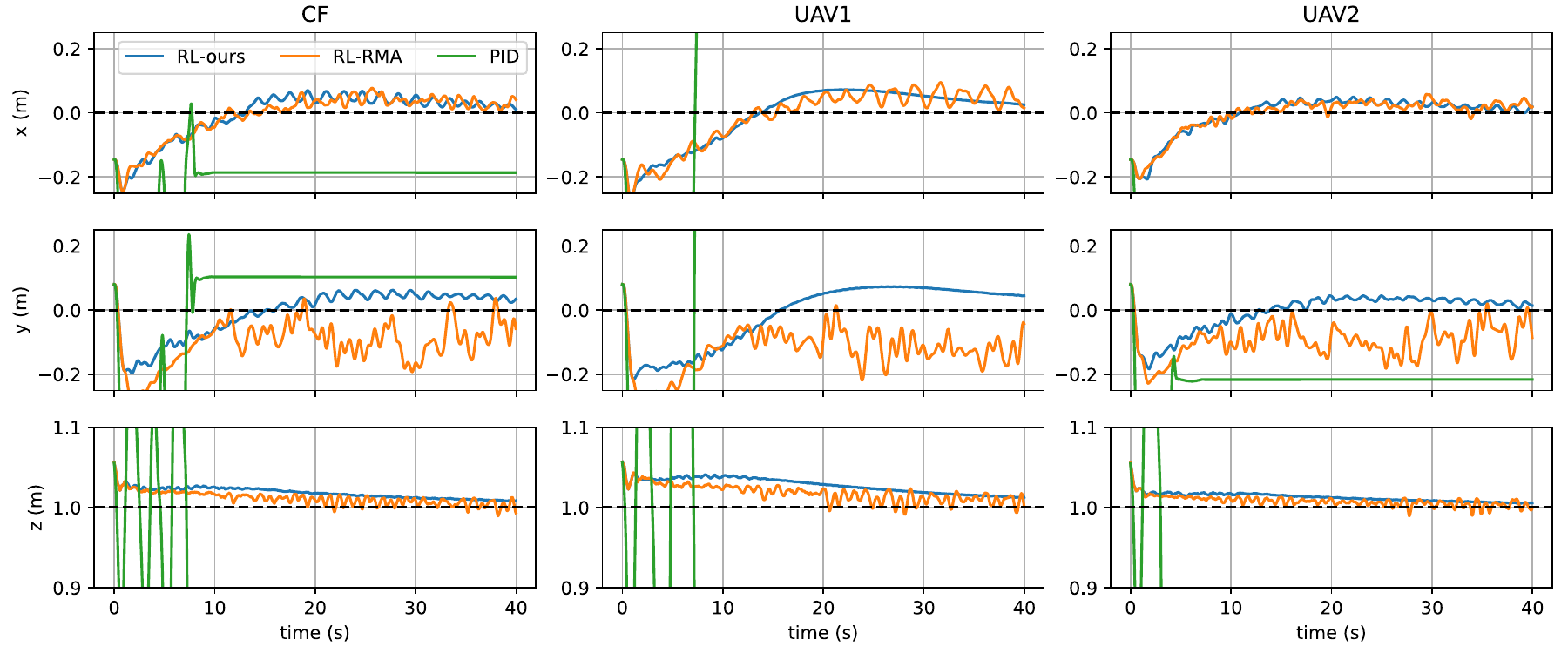}
    \caption{State comparison between our method and the baselines for each quadrotor under 30\% LoE of $F_2$. The black dashed lines are the target state.}
    \label{fig:state_comparison}
\end{figure}

We validate our framework in a simulation and compare it with two other methods.
\textbf{RL-ours} utilizes our hybrid controller and transformer-based adaptation module.
\textbf{RL-RMA}~\cite{kumar2021rma} shares the same controller but employs a 1-D CNN-based RMA-style adaptation module for benchmarking.
\textbf{PID} is a standard Proportional-Integral-Derivative controller without fault handling.
\textbf{No Fault (oracle)} evaluates performance without faults, representing the theoretical upper bound for RL-ours, which aims to match this level under optimal fault-tolerant control.

\begin{figure}[t]
    \centering
    \includegraphics[width=\columnwidth]{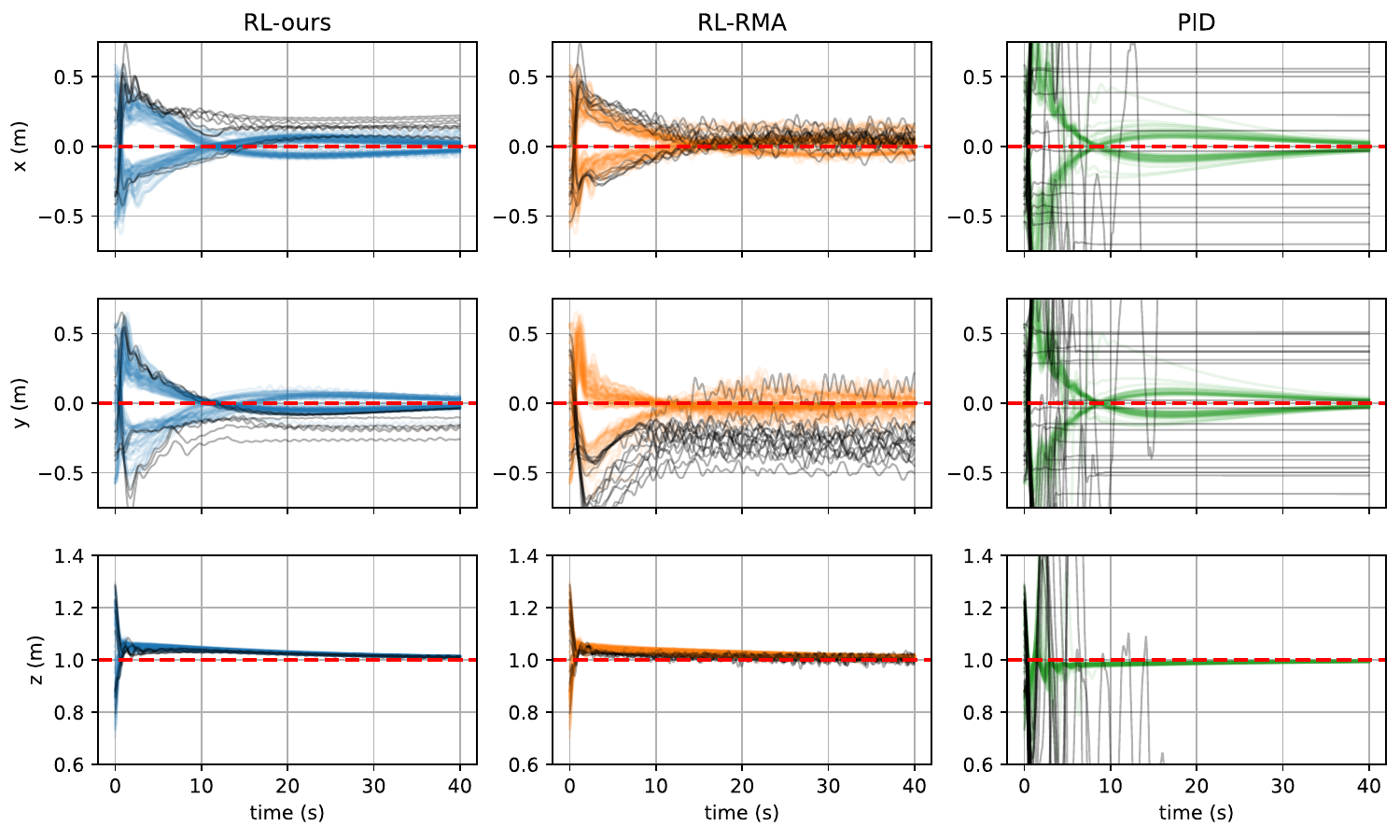}
    \caption{State comparison between our method and the baselines for the UAV1 model with 100 fault cases. Colored solid lines are successful cases, black solid lines are failure cases, and red dotted lines are the target state.}
    \label{fig:state_converge}
\end{figure}

\begin{figure}[t!]
    \centering
    \includegraphics[width=0.9\columnwidth]{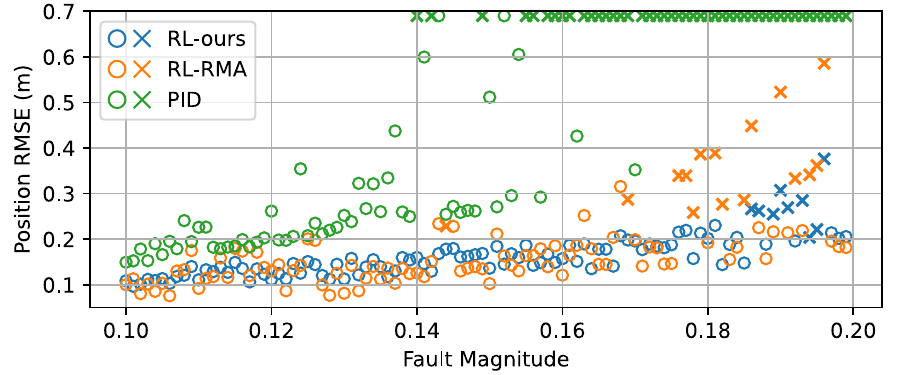}
    \caption{Position RMSE and success/failure results between our method and the baselines for the UAV1 model with 100 fault cases. The cases where the RMSE exceeds 0.7m are noted on the boundary.}
    \label{fig:winfail_100}
\end{figure}

\vspace{0.6em}\noindent\textbf{State Comparison:}\quad
We compare control performance under a 30\% thrust loss in motor 2 across three quadrotor models: CF, UAV1, and UAV2.
RL-ours demonstrates successful convergence within the margin, while RL-RMA shows significant oscillations and exceeds it.
PID becomes unstable and crashed after about 7 seconds.
Fig.~\ref{fig:state_comparison} shows state trajectories.

\vspace{0.6em}\noindent\textbf{Quantitative Testing:}\quad
We conduct quantitative tests on 100 random initial states, with positions and fault magnitudes sampled from Table~\ref{tab:cf2_parameters}.
Metrics include control success probability, root mean square error (RMSE) of position, and maximum position error.
Success is defined as maintaining a positional error within 0.173m for at least one second.
As shown in Table~\ref{tab:performance_and_std_dev}, our method outperforms others across all metrics, showing strong fault tolerance.
It achieves precise control with high success, even on untrained UAVs, demonstrating its adaptability.

\vspace{0.6em}\noindent\textbf{Robustness Testing:}\quad
We evaluate robustness by testing across the fault range using UAV1, an unseen model, with 100 test cases.
Fig.~\ref{fig:state_converge} shows state trajectories.
RL-ours stabilizes near targets, while the others often deviates significantly or oscillates.
Fig.~\ref{fig:winfail_100} demonstrates position RMSE and success status per case.
RL-ours maintains lower RMSE and higher success, even with significant LoE faults above 0.18.
The others struggle as fault magnitude increases.
This indicates our method's consistent fault tolerance, even in substantial faults and unseen dynamics.

\vspace{0.6em}\noindent\textbf{Latent Inference:}\quad
Finally, we compare the latent inference performance of the proposed transformer-based adaptation module with the RMA-style 1-D CNN.
RMSE $\|\hat z_t - z_t\|$ is measured between the adaptation module and the environmental factor encoder.
RL-ours achieves an RMSE of \mbox{$0.1113 \pm 0.0036$}, far lower than \mbox{$5.5798 \pm 0.0009$} of \mbox{RL-RMA}, demonstrating improved inference accuracy.
Notably, our method maintains precise latent inference, which confirms that our module effectively replaces the environmental encoder without privileged parameters.

\section{CONCLUSIONS}

We presented a reinforcement learning-based fault-tolerant control under uncertain and varying field conditions.
Inspired by recent advancements in RMA, our proposed framework enables robust real-time inference without the need for additional privileged information or retraining.
Experiments demonstrate that our method outperforms PID and CNN-based approaches across various actuator faults and unseen dynamic scenarios, including loss-of-effectiveness faults and variations in quadrotor physical parameters.
Our method maintains superior stability even beyond the training distribution.
These advantages make our framework reliable for practical field applications, even when the payload changes unexpectedly, thereby reducing mission complexity.

\addtolength{\textheight}{-12cm}   

\bibliographystyle{ieeetran}
\bibliography{IEEEabrv, bib}

\end{document}